\documentclass[letterpaper]{article} 
\usepackage[preprint]{aaai2027}  
\usepackage[hyphens]{url}  
\usepackage{graphicx} 
\usepackage{natbib}  
\usepackage{caption} 
\usepackage{algorithm}
\usepackage{algorithmic}

\usepackage{amsmath}

\usepackage{newfloat}
\usepackage{listings}
\DeclareCaptionStyle{ruled}{labelfont=normalfont,labelsep=colon,strut=off} 
\floatstyle{ruled}
\newfloat{listing}{tb}{lst}{}
\floatname{listing}{Listing}

\usepackage{booktabs}

\newcommand{\peng}[1]{{\color{blue}{#1}}}

\title{RL-Lock: Reinforcement Learning for Generating Interlocking Assemblies}
\author{
    Xuyang Ma\textsuperscript{\rm 1},
    Chaewoon Kim\textsuperscript{\rm 1},
    Haonan Zhang\textsuperscript{\rm 2},
    Rulin Chen\textsuperscript{\rm 3},
    Ziqi Wang\textsuperscript{\rm 2},
    Peng Song\textsuperscript{\rm 1}\corresponding
}
\affiliations{
    \textsuperscript{\rm 1}Singapore University of Technology and Design\\
    xuyang\_ma@mymail.sutd.edu.sg, chaewoon\_kim@mymail.sutd.edu.sg,\\
    peng\_song@sutd.edu.sg\\
    \textsuperscript{\rm 2}The Hong Kong University of Science and Technology\\
    haonan.zhang@connect.ust.hk, ziqiw@ust.hk\\
    \textsuperscript{\rm 3}Beijing Normal-Hong Kong Baptist University\\
    rulinchen@bnbu.edu.cn
}

\begin{document}

\maketitle


\begin{abstract}
An interlocking assembly is an assembly in which component parts are connected purely through their geometric arrangement, without relying on external connectors such as glue and nails.
Such assemblies have been widely used in a variety of real-world applications due to their structural stability.
The problem of generating interlocking assemblies is generally formulated as a shape decomposition problem, where a target 3D object represented as a voxel grid is partitioned into a prescribed number of interlocking pieces.
We observe that generating interlocking assemblies is inherently a sequential decision-making problem, where an agent repeatedly decides which piece each voxel should be assigned to. 
Inspired by the observation, we propose the first reinforcement learning framework {\em RL-Lock} for generating interlocking assemblies, without relying on handcrafted search heuristics as existing works did.
RL-Lock combines structured action chunking with MCTS-guided policy-value learning to efficiently navigate the large combinatorial search space for interlocking assembly generation.
We demonstrate through experiments that RL-Lock allows effective generation of interlocking assemblies, especially for challenging cases in which existing approaches take too long or even fail to find a valid solution.
\end{abstract}


\section{Introduction} 
\label{sec:intro}

\begin{figure}[!t]
	\includegraphics[width=0.99\columnwidth]{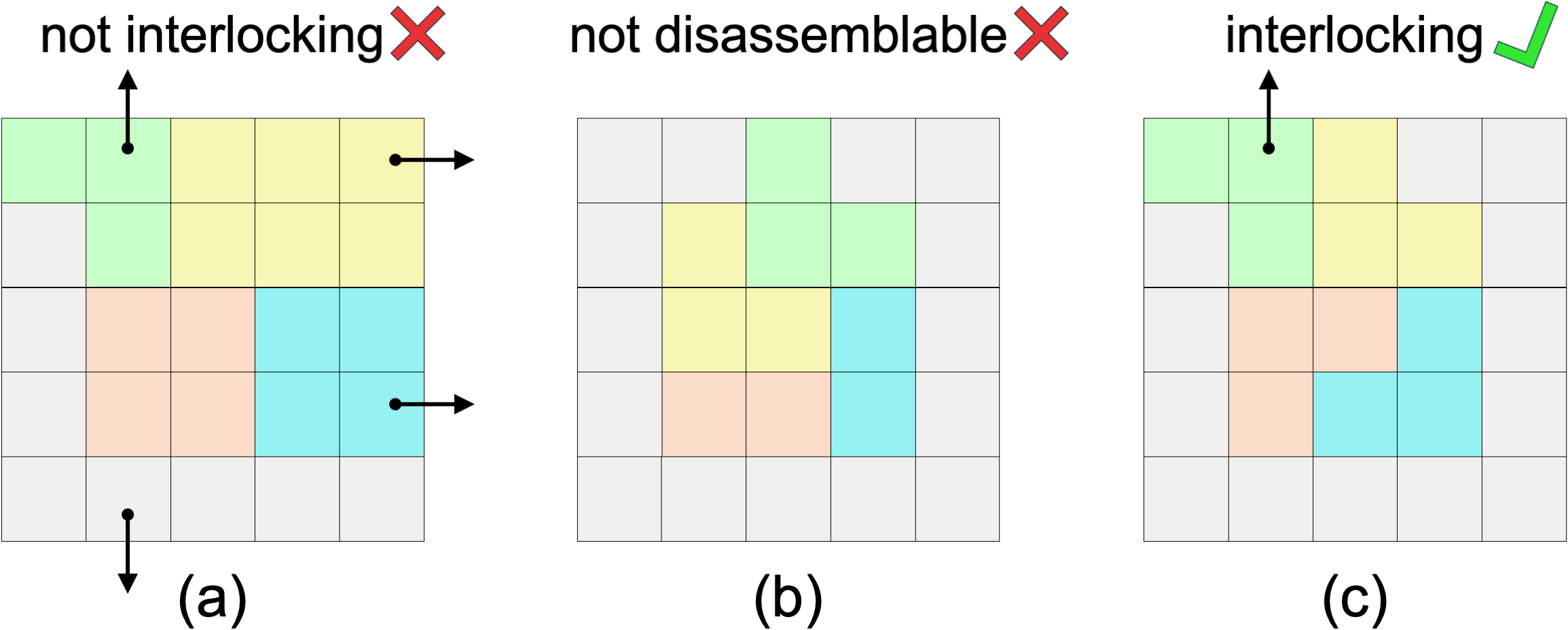}
	\caption{
	Taking a $5 \times 5$ square as an input shape, many decompositions will fail to generate an interlocking assembly such as those that are (a) not interlocking or (b) not disassemblable. 
	In this paper, we propose a reinforcement learning framework  to generate (c) interlocking assemblies with a single key (in green color). 	
	}
	\label{fig:interlocking_example}
\end{figure}

An interlocking assembly is an assembly in which component parts are connected through their geometric arrangement, making the assembly structurally stable without relying on external connectors such as glue and nails~\cite{Song-2022-InterlockingSurvey}.
In an interlocking assembly, parts need to follow certain orders to be assembled into the target object. 
Once assembled, there is only {\em one movable part}, called {\em the key}, while all other parts as well as any subset of parts are immobilized relative to one another; see Figure~\ref{fig:interlocking_example}(c) for a 2D example.
Such assemblies have been widely used in a variety of real-world applications, including 
puzzles~\cite{Song-2012-InterCubes, Chen-2022-HighLevelPuzzle}, 
furniture~\cite{Fu-2015-Furniture}, 
architectural structures~\cite{Wang-2019-TopoLock}, and
additive manufacturing~\cite{Yao-2017-InterlockShell}
since they ensure structural stability, simplify assembly, and facilitate disassembly.

The problem of generating interlocking assemblies is typically formulated as a shape decomposition problem. 
Given a target 3D object represented as a voxel grid, the objective is to decompose the object into multiple pieces whose union exactly reconstructs the input shape while simultaneously satisfying the interlocking property and disassemblability constraint.
Since the geometry of each piece is initially unknown, it has to be constructed by determining how voxels should be assigned to each piece such that the final decomposition forms a valid interlocking assembly. 
This coupling among geometric construction, interlocking property, and disassemblability constraint makes the problem highly challenging. Figure~\ref{fig:interlocking_example}(a\&b) shows two decompositions that fail to generate an interlocking assembly.


Existing approaches~\cite{Song-2012-InterCubes, Wang-2018-DESIA, Chen-2022-HighLevelPuzzle} address this problem primarily through combinatorial search.  
Starting from the input shape, these approaches iteratively construct the geometry of each piece by assigning voxels to it, evaluate whether the intermediate assembly satisfies the interlocking requirement, and continue searching until a valid decomposition is found. 
Since the search space grows exponentially with the number of voxels and pieces, all existing approaches rely on carefully designed heuristics to guide the search, including seed voxel selection and piece growth strategy. 
These heuristics significantly reduce computational complexity while improving the likelihood of finding feasible solutions.


Despite their success, heuristic-based approaches suffer from an inherent limitation: 
manually designed search strategies encode only limited prior knowledge and may overlook promising design solutions that are difficult for human experts to anticipate. 
As a result, these approaches often require excessive computation time and may even fail to generate interlocking assemblies for challenging cases involving a large number of pieces. 
This limitation motivates a fundamentally different approach that {\em learns to generate interlocking assemblies from experience} rather than relying on handcrafted search heuristics.


We observe that generating interlocking assemblies via shape decomposition is inherently a sequential decision-making problem. 
Taking a voxelized  shape as an input, an agent repeatedly decides which piece each voxel should be assigned to. 
Every decision changes the geometry of current piece and pieces to be constructed, and affects the future feasibility of the decomposition. 
The key challenge is that generating interlocking assemblies involves long-horizon sequential decision-making under {\em hard global geometric constraints} (i.e., interlocking and disassemblability), in which the consequences of each geometric decision may not become apparent until the entire assembly has been constructed.


To address this challenge, we formulate interlocking assembly generation as a reinforcement learning (RL) problem, in which the state encodes the current decomposition, the action progressively constructs each piece by assigning voxels, and the reward function jointly captures geometric validity, interlocking, and disassemblability. 
To reduce the effective decision horizon, we introduce structured action chunking, which groups a sequence of low-level actions (i.e., voxel assignments) into a single high-level action corresponding to the construction of a piece. 
We further propose RL-Lock, an RL framework that trains a policy-value neural network via Monte Carlo tree search (MCTS) to guide the selection of high-level actions for interlocking assembly generation. 
By learning from previous search episodes, the policy and value functions capture patterns associated with successful decompositions, enabling MCTS to prioritize promising branches and efficiently navigate the large combinatorial search space.


Our main contributions are summarized as follows:

\begin{itemize}
	\item 
	We propose the first reinforcement learning framework for generating interlocking assemblies without relying on handcrafted search heuristics, and evaluate its performance by comparing it with two baseline approaches.
	
	\item 
	We demonstrate through experiments that RL-Lock allows effective generation of interlocking assemblies, especially for challenging cases that existing approaches take too long or even fail to find a valid solution.
\end{itemize}


\section{Related Work}
\label{sec:relatedwork}


\paragraph{Reinforcement learning for design generation.} 
Reinforcement learning was originally developed to solve sequential decision-making problems under delayed rewards and unknown environments. 
Recent advances have extended its application beyond games and control to design generation tasks, including
game level design~\cite{Khalifa-2020-PCGRL, Earle-2025-GameLevelDesign},
procedural material design~\cite{Li-2024-ProceduralMaterial},
CAD model generation~\cite{Li-2026-ReCAD, Niu-2026-FromIntentToExecution},
compliant mechanism design~\cite{Choi-2025-CompliantMechanism}, and
modular robot design synthesis~\cite{Whitman-2020-ModularRobot}.

In particular, reinforcement learning methods have been developed to generate designs of assemblies, which can be classified into two classes. 
The first class uses a bottom-up strategy to generate designs of assemblies via selecting and assembling the instances of a prescribed set of building blocks.
Hosmer et al.~\shortcite{Hosmer-2020-SpatialAssembly} generated architectural assemblies from sets of digitally encoded spatial parts designed by architects and assembled using policies learned through self-play reinforcement learning.
Chung et al.~\shortcite{Chung-2021-BrickByBrick} designed an RL agent to assemble unit primitives (i.e., LEGO bricks) sequentially to approximate desired targets represented by 2D images.
Wibranek et al.~\shortcite{Wibranek-2021-SLBlocksAssembly} proposed a reinforcement learning approach for sequential assembly of just one type of element called SL block~\cite{Shih-2016-SLBlocks} to approximate a prescribed 2D/3D curve.
Vallat et al.~\shortcite{Vallat-2023-Scaffold-freeConstruction} used multi-agent reinforcement learning for the design of spanning structures from two types of blocks (i.e., hexagonal block and hourglass-shaped block) such that multiple robot arms can construct the structures without the need for scaffolding.
The second class uses a top-down strategy to generate assembly designs by decomposing a target 3D shape into a set of parts and, to the best of our knowledge, is represented by only a single work.
Seriket et al.~\shortcite{Seriket-2025-Multi-materialPrinting} developed a Q-learning algorithm to decompose a voxelized 3D shape into a set of parts for multi-material additive manufacturing, where each part can move along a single direction in the resulting assembly. 

In this work, we focus on top-down generation of interlocking assemblies~\cite{Song-2022-InterlockingSurvey}.
Compared with assemblies designed in~\cite{Seriket-2025-Multi-materialPrinting}, interlocking assemblies are much more difficult to design due to the global property of single-key interlocking.
To address this challenging problem, we propose the first reinforcement learning framework that employs Monte Carlo tree search with action chunking, guided by a policy-value neural network.

\begin{figure*}[!t]
	\centering
	\includegraphics[width=2.08\columnwidth]{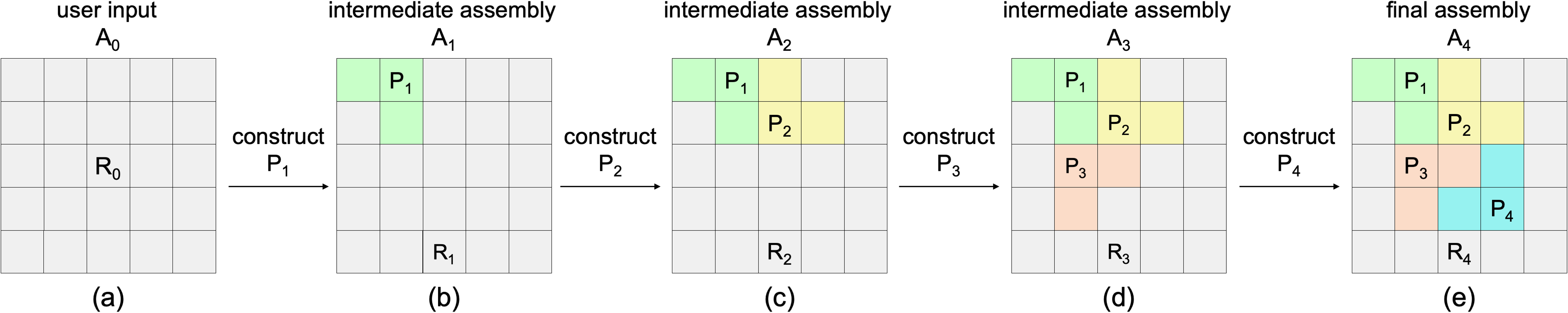}
	\caption{
		Illustration of the iterative shape decomposition via a 2D example.
		Taking (a) a 2D shape as an input,  we generate (e) an interlocking assembly by iteratively constructing the geometry of each piece $P_i$ via assigning voxels from $R_{i-1}$ to $P_i$ while ensuring (b-e) each intermediate assembly $A_i$ is interlocking and disassemblable. 	
	}
	\label{fig:construction_framework}
\end{figure*}


\paragraph{Design of interlocking assemblies.} 
Interlocking assemblies are intriguing since they are structurally stable purely based on their component parts' geometric arrangement~\cite{Song-2022-InterlockingSurvey}.
However, designing interlocking assemblies is extremely hard for humans, even for skilled professional,  since every part (except the key) as well as any subset of parts should be immobilized by their geometric arrangement.
As mentioned in~\cite{Coffin-1990-Puzzling}, designing interlocking assemblies requires hours, or even days, of mental work. 

Since 2010s, a few heuristic-based approaches have been proposed for computational design of 3D interlocking assemblies.
When a target shape is given, designing interlocking assemblies is formulated as a shape decomposition problem, e.g., to design 
interlocking puzzles for recreation~\cite{Xin-2011-BurrPuzzles, Song-2012-InterCubes, Tang-2019-3DDissection, Chen-2022-HighLevelPuzzle} or 
interlocking assemblies for 3D printing large objects~\cite{Song-2015-Interlock}.
When a set of initial parts without joints is given, designing interlocking assemblies is formulated as a joint planning problem, e.g., for designing interlocking furniture~\cite{Fu-2015-Furniture, Song-2017-ReconfigInterlock} or interlocking shells for 3D printing~\cite{Yao-2017-InterlockShell}.
In particular, Wang et al.~\shortcite{Wang-2018-DESIA} developed a unified framework to design interlocking assemblies of different forms by leveraging a graph-based representation.

In this paper, we also formulate designing interlocking assemblies as a shape decomposition problem. 
Rather than relying on handcrafted search heuristics used by existing approaches, we investigate a fundamentally different paradigm in which a reinforcement learning agent learns an effective design policy directly from experience.
Experiments demonstrate that our RL approach allows effective generation of interlocking assemblies, even for challenging cases that existing approaches struggle to find a valid solution.

\section{Problem Formulation}
\label{sec:problem}

To design interlocking assemblies, the user inputs are a 3D shape represented by a voxel grid with resolution $W \times H \times D$, and a desired number of pieces, denoted by $K$, in the resulting assembly.
Given the user inputs, our goal is to design interlocking assemblies by decomposing the input voxelized shape into $K$ interlocking pieces, where all the pieces, except the last one, should have the same number of voxels denoted by $n$.
In our experiments, $n$ is set as $\lfloor N / K \rfloor$ by default, where $N$ is the total number of voxels in the voxelized shape.

\paragraph{Iterative shape decomposition.}  \
Given the input shape denoted as $R_0$, we perform iterative shape decomposition to construct the geometry of each piece one by one, following the approach in~\cite{Wang-2018-DESIA}. 
This forms a sequence of constructed pieces, $P_1$, $P_2$, $...$, $P_{K-1}$, with $R_{K-1}$, the remaining volume of $R_0$, as the last piece:
\begin{displaymath}
	[R_0] \rightarrow [P_1,R_1] \rightarrow [P_1,P_2,R_2] \rightarrow ... \rightarrow [P_1,...P_{K-1},R_{K-1}] \ .
\end{displaymath}
Here we denote each intermediate assembly $[P_1, ..., P_i, R_i]$ as $A_{i}$ ($0 \leq i \leq K-1$);
see Figure~\ref{fig:construction_framework}.  

To guarantee that the resulting assembly $\mathbf{A}_{K-1}$ is interlocking and disassemblable, we have the following requirements when decomposing $R_{i-1}$ into $P_i$ and $R_i$:
\begin{enumerate}
	\item
	{\em  Connected.} \
	The geometries of $P_i$ and $R_i$ should each be connected, making $A_i$ a valid assembly. 
	
	\item
	{\em Interlocking.} \
	$A_i$ ($i \geq 2$) is interlocking with $P_1$ as the key, which is the only movable piece in the assembly.

	\item
	{\em Removable.} \
	$P_i$ can be removed from $[P_i, R_i]$, so we can disassemble $A_i$ in the order of $P_1$, $P_2$,  $...$,  $P_i$, $R_i$.
\end{enumerate}	

The advantage of this iterative shape decomposition framework is that we achieve the goal of global interlocking by satisfying a set of local requirements when constructing each pair of $P_i$ and $R_i$.

\paragraph{Finite Markov Decision Process.}
We observe that generating interlocking assemblies via iterative shape decomposition is inherently a sequential decision-making problem, where an agent repeatedly decides which voxel should be assigned from the remaining volume $R_{i-1}$ in $A_{i-1}$ to the current piece $P_i$ in $A_i$.

We model this sequential decision-making problem as a Finite Markov Decision Process (MDP), and describe the core components of the MDP as below.
\begin{enumerate}
	\item
	{\em Actions.} \
	An action assigns one voxel from $R_{i-1}$ to $P_i$ before the number of voxels in $P_i$ reaches the target $n$; see  Figure~\ref{fig:action_chunking}(a\&c) for an example.
	
	\item
	{\em  States.} \
	A state is an intermediate assembly via decomposition of the input shape $R_0$.
	
	\item
	{\em Transition Probabilities.} \
	Once an action is performed, the current state transitions deterministically to the next state, meaning that the transition probability is always 1.
	
	\item
	{\em Rewards.} \
	The action receives rewards if the decomposition satisfies the three requirements above.
\end{enumerate}	





\section{Method}
\label{sec:method}

We observe that solving the MDP requires long-horizon decision-making, as generating a complete interlocking assembly involves a long sequence of voxel assignment actions, where the number of actions is $n(K-1) \approx N$.
To reduce the difficulty of long-horizon decision making, we propose to perform structured action chunking by grouping a sequence actions for constructing each piece $P_i$ as an action chunk, resulting in $K-1$ action chunks (Section~\ref{subsec:action_chunking}).
Based on the action chunking, we propose a reinforcement learning approach to solve the above MDP problem (Section~\ref{subsec:RL_framework}).

\subsection{Action Chunking}
\label{subsec:action_chunking}

\begin{figure}[!t]
	\includegraphics[width=0.99\columnwidth]{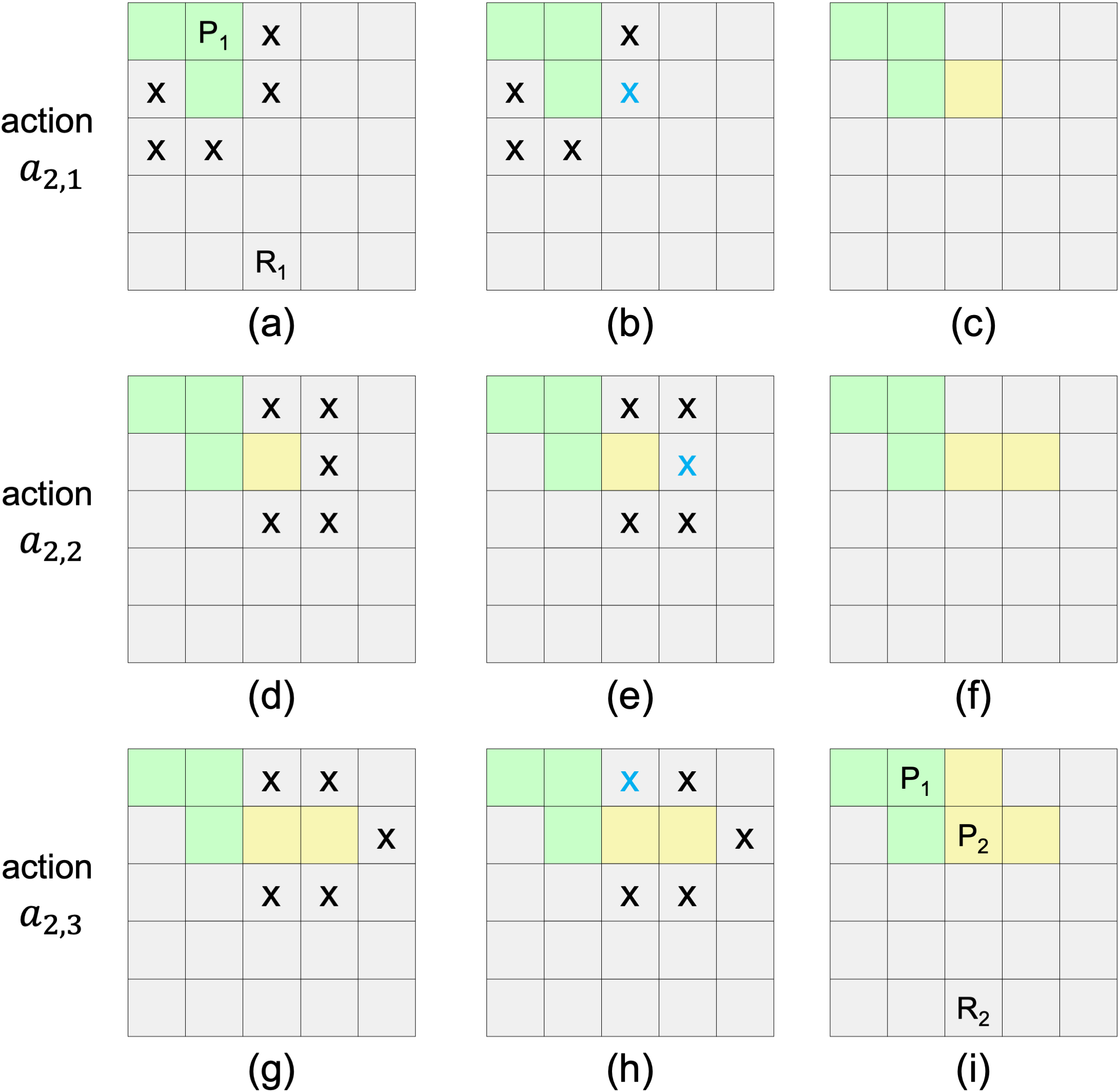}
	\caption{
		An action chunk $c_2$ (consisting of 3 actions) to construct the piece $P_2$'s geometry.
		To assign the first voxel, the action $a_{2, 1}$ 
		(a) identifies the legal frontier (highlighted with a black X) for the voxel assignment, 
		(b) selects one voxel (highlighted with a blue X) from the candidates and 
		(c) assigns it from $R_1$ to $P_2$.
		The actions $a_{2, 2}$ and $a_{2, 3}$ are performed following the same procedure.
	}
	\label{fig:action_chunking}
\end{figure}

We propose to perform action chunking by grouping a sequence of low-level actions (i.e., voxel assignments) into a single high-level action (i.e., piece geometry construction); see Figure~\ref{fig:action_chunking} for an example.

\paragraph{Action chunking.}
At an intermediate assembly $A_{i-1}$, we define an action chunk 
\begin{displaymath}
c_i = (a_{i,1}, \ a_{i,2},  \  \ldots,  \ a_{i, n})
\end{displaymath}
as an ordered sequence of actions $\{ a_{i, j} \}$, $j \in [1, n]$, where each action assigns one voxel from the remaining volume $R_{i-1}$ to the current piece $P_i$ for constructing $P_i$'s geometry.

To facilitate interlocking, the first voxel of $P_i$ is sampled from the six-connected frontier of $P_{i-1}$; see Figure~\ref{fig:action_chunking}(a-c) for a 2D example.
To ensure a connected geometry for piece $P_i$, each subsequent voxel is sampled from the six-connected frontier of the current geometry of $P_i$; see Figure~\ref{fig:action_chunking}(d-f) and (g-i) for two 2D examples.

\paragraph{State checkers.}
By action chunking, each state $s_i$ is limited to an intermediate assembly $A_i$, where each piece has been completely constructed; see Figure~\ref{fig:action_chunking}(i) for an example.
Given a state $s_i$, we propose three state checkers to test if it satisfies the three requirements in Section~\ref{sec:problem}.
\begin{enumerate}
	\item
	{\em  Connectivity checker.} \
	The connectivity of each piece $P_i$'s geometry is ensured via the piece construction procedure; see again Figure~\ref{fig:action_chunking}.
	Hence, we only need to check the connectivity of the remaining volume $R_i$.
	To this end, we perform a breadth-first search (BFS) over the remaining volume $R_i$'s voxel connection graph. 
	$R_i$ is connected if all its voxels have been visited by the BFS.
	

	\item
	{\em Interlocking checker.} \
	To check whether a state $s_i$ is interlocking, we use the graph-based testing approach in~\cite{Wang-2018-DESIA}, which first builds base directional blocking graphs for the intermediate assembly  and then checks if these graphs are strongly connected except the key piece.  
	
	\item
	{\em Removability checker.} \
	To check removability of the piece $P_i$ from $[P_i, R_i]$, we test if the remaining volume $R_i$ blocks the piece $P_i$'s translational movement along each of 6 axial directions in 3D. 
	$P_i$ is removable from $[P_i, R_i]$ if $P_i$ is not blocked by $R_i$ in at least one axial direction.
\end{enumerate}

\begin{figure}[!t]
	\includegraphics[width=0.99\columnwidth]{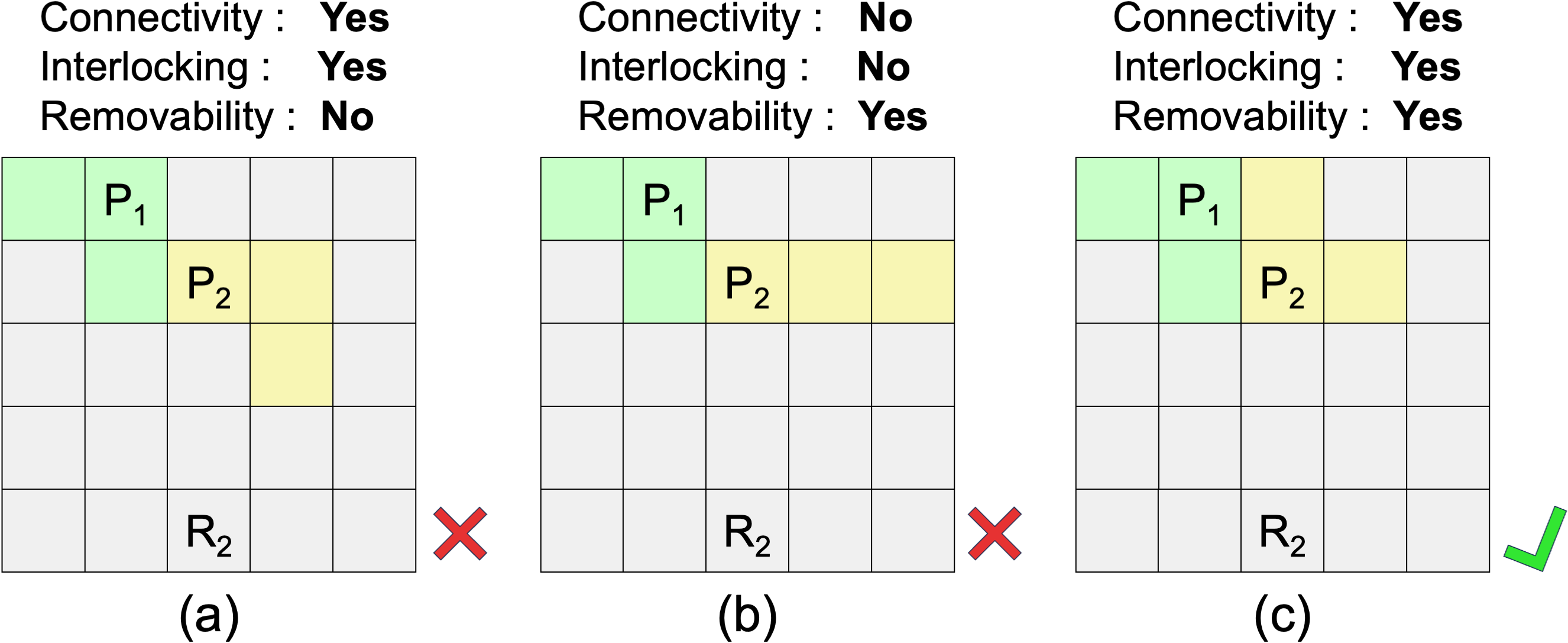}
	\caption{
		Three example states of $\mathbf{A}_2$. 
		(a) The state is invalid since $P_2$ is not removable in $[P_2, R_2]$.
		(b) The state is invalid since $R_2$ is disconnected (with two connected components) and both $P_1$ and $P_2$ are movable in $\mathbf{A}_2$.
		(c) The state is valid since it passes all the three checkers. 
	}
	\label{fig:state_checker}
\end{figure}

\subsection{Reinforcement Learning}
\label{subsec:RL_framework}

We observe that generating interlocking assemblies via iterative shape decomposition shares some similarities with playing the game of Go.
First, the states of interlocking assemblies and the game of Go are both represented using a regular grid. 
Second, the action chunk of constructing the geometry of a piece is similar to the action of moving a stone in the game of Go.
Lastly, the success of generating interlocking assemblies can only be determined at the end, once all the pieces have been constructed, similar to playing the game of Go.
Inspired by this observation, we propose a reinforcement learning approach for generating interlocking assemblies, following a paradigm pioneered by AlphaGo and subsequent works~\cite{Silver-2016-AlphaGo, Silver-2017-AlphaGoZero}.

\begin{figure}[!t]
	\includegraphics[width=0.99\columnwidth]{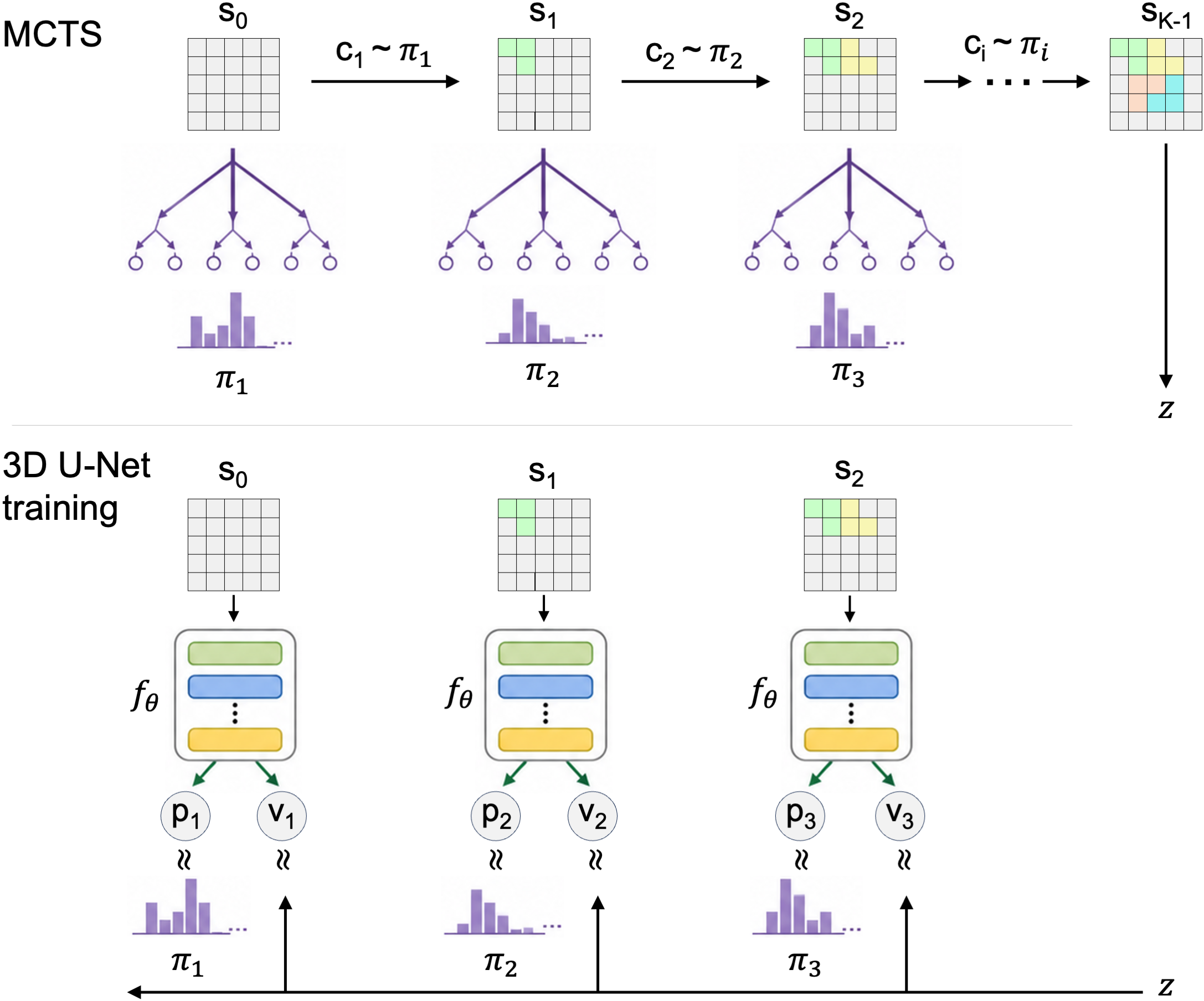}
	\caption{
		Our reinforcement learning framework to generate interlocking assemblies. 
		In each state $s_{i-1}$, an MCTS is executed guided by the latest neural network $f_\theta$.
		Action chunks are selected according to the search probabilities computed by the MCTS, $c_i \sim \boldsymbol{\pi}_i$.
		During training, the neural network parameters $\theta$ are updated to maximize the similarity of the policy vector $\mathbf{p}_i$ to the search probabilities $\boldsymbol{\pi}_i$, and to minimize the error between the value $v_i$ and the cumulative reward $z_i$.
	}
	\label{fig:RL_framework}
\end{figure}

\paragraph{RL framework.}
Our RL framework employs Monte Carlo tree search (MCTS) guided by a neural network; see Figure~\ref{fig:RL_framework}. 
Our framework uses a policy--value neural network $f_\theta$ with parameters $\theta$.
The neural network takes the state $s$ of an intermediate assembly as an input, and outputs both probabilities $\mathbf{p}$ of selecting action chunks from that state and a value $v$, i.e., $(\mathbf{p}, v) = f_\theta(s)$.
The vector of probabilities $\mathbf{p}$ represents the probability of selecting each action chunk $c$ while the value $v$ is a scalar evaluation that estimates the expected cumulative reward from state $s$.
We implement the neural network using a 3D U-Net architecture~\cite{Cicek-2016-3DUNet}.

To train the neural network, in each state $s$, an MCTS search is executed, guided by the neural network $f_\theta$. 
The MCTS outputs probabilities $\boldsymbol{\pi}$ of selecting each action chunk, which is usually better than the raw selection probabilities $\mathbf{p}$ of the neural network $f_\theta$.
Therefore, our framework uses these search operators repeatedly in a policy iteration procedure.
The neural network's parameters are updated to make the action chunk selection probabilities and value $(\mathbf{p}, v) = f_\theta(s)$ more closely match the improved search probabilities and cumulative reward $(\boldsymbol{\pi}, z)$.
These new parameters are used in the next iteration of MCTS to make the search even stronger. 

\paragraph{MCTS.}
In the MCTS tree, each node represents a state $s_{i}$, and each edge selects and applies one candidate action chunk \(c_{i+1}\) to produce the next state $s_{i+1}$; see Figure~\ref{fig:RL_framework} (top).
Given a state $s_{i}$, we generate \(M_s\) raw action chunks by sampling each voxel uniformly from the current legal frontier.
We discard checker-rejected action chunks and sample $M_c$ chunks from the remaining valid pool to form the candidate set of action chunks for MCTS.
Since candidate action chunks are checker-accepted, each valid transition advances the decomposition by one piece and receives a reward of \(1/K\). 
A search branch starts from the root state $s_0$ (i.e., the user input $R_0$), and terminates
unsuccessfully when no checker-accepted continuation can be found or 
successfully when \(K-1\) action chunks have been accepted.

\paragraph{Training.}
We train the policy--value neural network using trajectories generated by MCTS.
At each state \(s_{i}\), MCTS uses the current neural network $f_\theta(s_{i})$ 
to guide the search over the candidate action chunks and returns probabilities $\boldsymbol{\pi}$ over them. 
After selecting an action chunk \(c_{i+1}\), the trajectory records the state \(s_{i}\), 
its candidate set of action chunks, the MCTS probability distribution
\(\boldsymbol{\pi}\), the immediate reward and terminal signal produced by this transition. 
Once the episode terminates, for each state \(s_{i}\), we sum the rewards obtained from that assembly until termination and use this cumulative reward $z_{i}$ as its value target. 
We then add \(s_{i}\), its candidate set of action chunks, MCTS probability distribution \(\boldsymbol{\pi}\), and cumulative reward $z_{i}$ to the replay buffer.

We sample mini-batches from the replay buffer and jointly optimize the entire network using
$$
\mathcal{L}(\theta)
=
-\boldsymbol{\pi}^T \log{\mathbf{p}}
+
(z-v)^2,
$$
where the first term is the cross-entropy between the MCTS action weights \(\boldsymbol{\pi}\) and the network probabilities
\(\mathbf{p}\), and the second term is the squared error between the cumulative reward \(z\) and the predicted value \(v\).

\section{Experiments}
\label{sec:experiments}


\begin{figure}[!t]
	\centering
	\includegraphics[width=0.99\columnwidth]{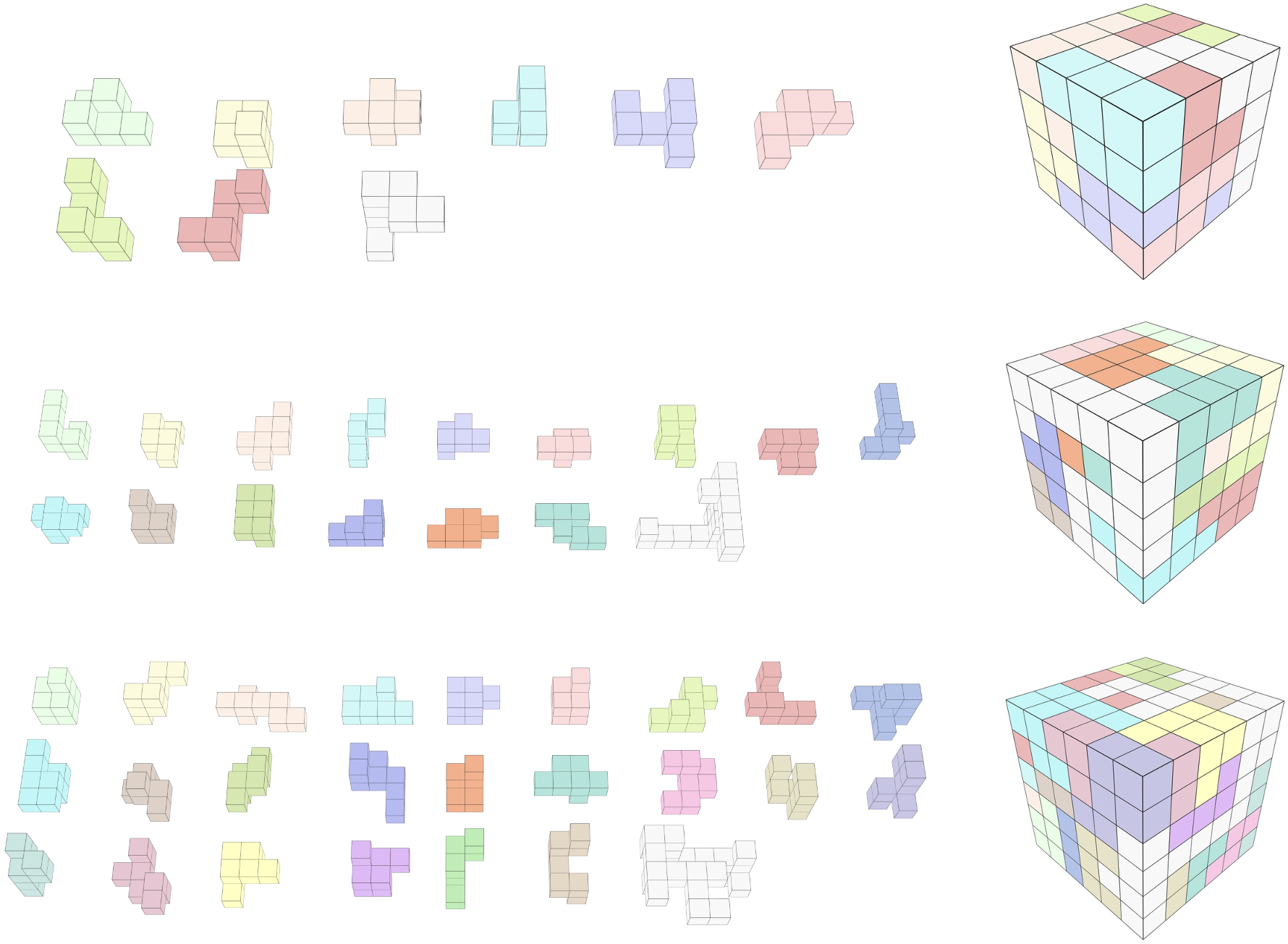}
	\caption{
		Representative interlocking cubes generated by RL-Lock:
		(top) 9-piece  \(4^3\) cube, 
		(middle) 16-piece \(5^3\) cube, and 
		(bottom) 25-piece \(6^3\) cube.
		Each row shows the generated pieces on the left and the assembled cube on the right.
	}
	\label{fig:result_cubes}
\end{figure}

\paragraph{Implementation.}
We implemented RL-Lock in JAX~\cite{Bradbury-2018-JAX} and conduct all experiments on a Linux workstation equipped with an AMD Ryzen 5 7500F CPU, 32 GB of RAM, and an NVIDIA GeForce RTX 5090 GPU.

All RL methods use a fixed \(8\times8\times8\) canvas, the smallest cubic canvas
that accommodates all evaluated inputs; smaller input domains are padded to this size.
For RL-Lock, at each state we sample up to $M_s=1024$ raw action chunks and draw $M_c=32$ candidates from those accepted by the state checkers.
MCTS performs 256 simulations per decision and considers at most 16 of these candidates.
We train the policy--value network with Adam at a learning rate of $5\times10^{-5}$, initialize it with random seed 0, and report the checkpoint with the best validation performance for each task.

\paragraph{Ablation study.}
In this ablation study, we compare RL-Lock with two baselines on the tasks of generating interlocking assemblies by decomposing cubes of different resolutions into a target number of pieces.
The purpose is to validate the usefulness of two key components in our framework, i.e., action chunking and MCTS.

{\em Generation tasks.} \
In this study, we have nine generation tasks via decomposing cubes of three different resolutions (i.e., $4^3$, $5^3$, and $6^3$ cubes), where we have three target piece counts per cube resolution; see Table~\ref{tab:overall-performance} (left).
Note that for each cube resolution, we choose a relatively large $K$ to make the task challenging.

{\em Two baselines.} \
We compare RL-Lock with two RL baselines for addressing the nine generation tasks.
\begin{enumerate}
	\item 
	\emph{RL-Lock\_Action.} \
	RL-Lock\_Action is a modified version of RL-Lock, where the key difference is that it selects one voxel per action without performing action chunking.
	We use 128 simulations per voxel decision, a maximum search depth of 32, and at
	most 16 considered actions.
	
	\item 
	\emph{PPO.} \
	Compared with RL-Lock\_Action, this baseline uses Proximal Policy Optimization (PPO)~\cite{Schulman-2017-PPO} rather than MCTS to learn a policy from experience.
	We set the discount factor to \(\gamma=1.0\), matching the undiscounted cumulative checker
	reward used by MCTS. 
	At inference, actions are sampled from the masked policy distribution at temperature 1.0.
\end{enumerate}
All the three methods use the same state checkers, normalized per-piece reward, fixed voxel canvas, and 3D U-Net backbone.
In addition, the two baselines share the same action spaces and legal-action mask rules.
We evaluate checkpoints on 64 validation episodes, stop training when the training losses stabilize, and use the best-performing checkpoint for final evaluation.

\begin{table}[t]
	\centering
	\footnotesize
	\setlength{\tabcolsep}{1.5pt}
	\begin{tabular*}{\columnwidth}{@{\extracolsep{\fill}}cc *{6}{c}@{}}
		\toprule
		\multicolumn{2}{c}{Input} & \multicolumn{2}{c}{PPO}
		& \multicolumn{2}{c}{RL-Lock\_Action}
		& \multicolumn{2}{c}{RL-Lock} \\
		\cmidrule(lr){1-2}
		\cmidrule(lr){3-4}
		\cmidrule(lr){5-6}
		\cmidrule(lr){7-8}
			Cube & \(K\) & Train & Infer.
			& Train & Infer.
			& Train & Infer. \\
			\midrule
			\(4^3\) & 7 & 29.40 & \textbackslash & 62.40 & \textbf{0.01} & 33.00 & 0.05 \\
			\(4^3\) & 8 & 21.60 & \textbackslash & 83.40 & \textbackslash & 49.20 & \textbf{0.47} \\
			\(4^3\) & 9 & 29.40 & \textbackslash & 114.00 & \textbackslash & 35.40 & \textbf{1.22} \\
			\midrule
			\(5^3\) & 13 & 76.80 & \textbackslash & 165.60 & \textbf{0.01} & 66.60 & 0.60 \\
			\(5^3\) & 14 & 85.20 & \textbackslash & 132.00 & \textbackslash & 80.40 & \textbf{0.84} \\
			\(5^3\) & 16 & 76.20 & \textbackslash & 148.80 & \textbackslash & 94.80 & \textbf{1.68} \\
			\midrule
			\(6^3\) & 20 & 59.40 & \textbackslash & 144.60 & \textbackslash & 63.00 & \textbf{2.86} \\
			\(6^3\) & 22 & 91.20 & \textbackslash & 195.00 & \textbackslash & 74.40 & \textbf{3.46} \\
			\(6^3\) & 25 & 107.40 & \textbackslash & 100.20 & \textbackslash & 90.60 & \textbf{4.87} \\
			\bottomrule
		\end{tabular*}
		\caption{Ablation study by comparing RL-Lock with two RL baselines in nine tasks of generating interlocking cubes with different resolutions and $K$s.
		Train denotes the wall-clock training time until the training losses stabilize.		
		Infer. is the accumulated inference time divided by the number of valid assemblies found.
		Both time measurements are reported in minutes.
		\textbackslash{} indicates that no valid solution was found within 60 minutes.
			}
	\label{tab:overall-performance}
\end{table}

{\em Statistics.} \
For each method, we report the wall-clock training time until the training
losses stabilize.
During inference, we evaluate 64 episodes per batch until either at least five valid assemblies are found or the total inference time reaches 60 minutes. 
We then compute the average inference time per valid assembly by dividing the accumulated inference time by the number of valid assemblies found.
Table~\ref{tab:overall-performance} summarizes these timing statistics for each task.

Table~\ref{tab:overall-performance} shows that RL-Lock succeeded on all the nine tasks while RL-Lock\_Action only succeeded on the \((4^3,K=7)\) and \((5^3,K=13)\) tasks.
The broader solution coverage of RL-Lock suggests that action chunking improves scalability to harder tasks since it significantly reduces the number of decisions to make from $n(K-1)$ (RL-Lock\_Action) to $K-1$ (RL-Lock).
In addition, we observed that RL-Lock's inference time increases with \(K\) for a fixed cube resolution.
This is expected since generating interlocking assemblies with a larger $K$ requires more decisions to make.

Table~\ref{tab:overall-performance} also shows that PPO did not find a valid solution within the 60-minute limit for any of the nine tasks.
Note that the only difference between PPO and RL-Lock\_Action is the strategy to learn the policy for assigning voxels. 
The performance difference suggests that MCTS enables the voxel-level approach to find valid solutions whereas PPO does not, supporting our choice of MCTS-guided learning over PPO.


\begin{table}[t]
	\centering
	\footnotesize
	\setlength{\tabcolsep}{1.5pt}
	\begin{tabular*}{0.75\columnwidth}{@{\extracolsep{\fill}}cccc@{}}
		\toprule
		\multicolumn{2}{c}{Input} 
		& \multicolumn{1}{c}{DESIA}
		& \multicolumn{1}{c}{RL-Lock} \\
		\cmidrule(lr){1-2}
		\cmidrule(lr){3-3}
		\cmidrule(lr){4-4}
		Cube & \(K\) & Comp. 
		 & Infer. \\
		\midrule
		$4^3$     &   7   & \textbf{0.01} & 0.05 \\
		$4^3$     &   8   & 0.50 & \textbf{0.47} \\
		$4^3$    &   9   & \textbackslash & \textbf{1.22} \\
		\midrule
		$5^3$    & 13   & \textbf{0.01} & 0.60 \\
		$5^3$    & 14   & 0.86 & \textbf{0.84} \\
		$5^3$    & 16   & 21.36 &  \textbf{1.68} \\
		\midrule
		$6^3$    & 20   & \textbf{0.12} & 2.86 \\
		$6^3$   & 22   & \textbf{2.54} &  3.46 \\
		$6^3$   & 25   & 32.26 &  \textbf{4.87} \\
		\bottomrule
	\end{tabular*}
	\caption{Comparison with DESIA on the nine generation tasks. 
		 DESIA is run repeatedly until five valid solutions are generated while RL-Lock follows the batched inference protocol in Table~\ref{tab:overall-performance}.
		 Comp. is total computation time of DESIA divided by five.
		 Infer. is the average inference time of RL-Lock.
		 Both time measurements are reported in minutes.
		 \textbackslash{} indicates that no valid solution was found within 60 minutes.
	}
	\label{tab:desia-comparison}
\end{table}

\begin{figure*}[!t]
	\centering
	\includegraphics[width=2.08\columnwidth]{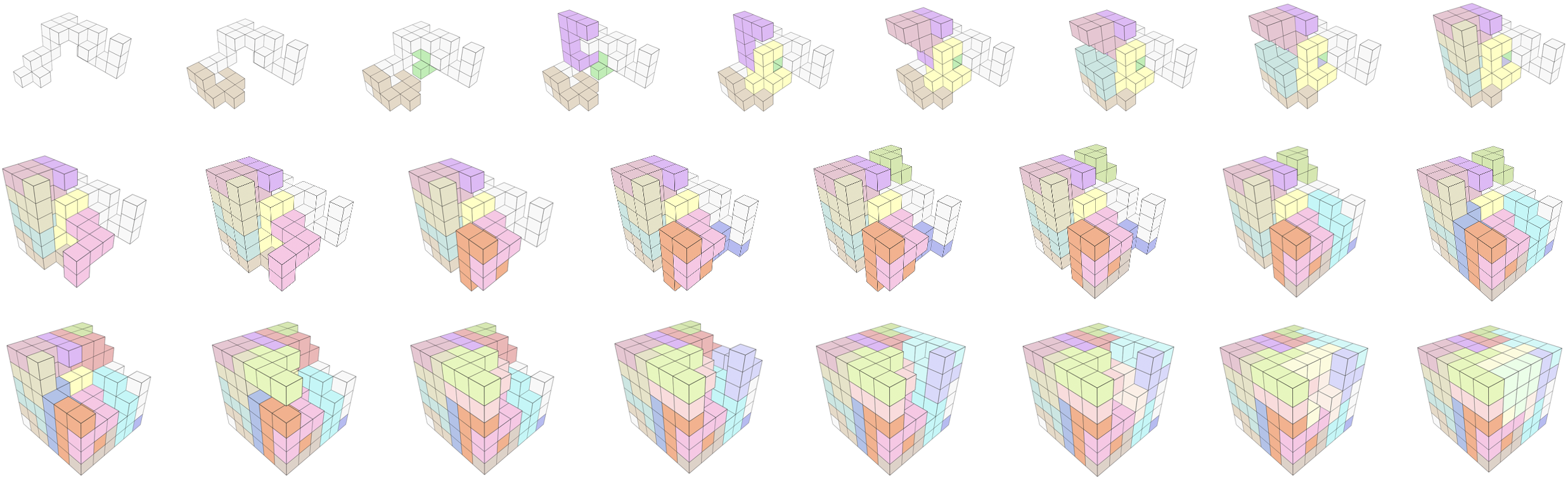}
	\caption{
		Assembly of a 25-piece $6 \times 6 \times 6$ interlocking cube generated by RL-Lock.
	}
	\label{fig:result_assembly_cube}
\end{figure*}

\begin{figure*}[!t]
	\centering
	\includegraphics[width=2.08\columnwidth]{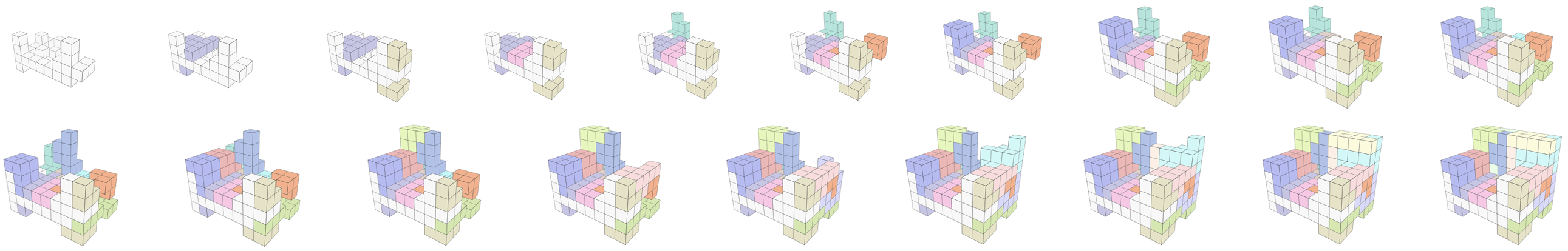}
	\caption{
		Assembly of a 19-piece interlocking Sofa generated by RL-Lock.
	}
	\label{fig:result_assembly_sofa}
\end{figure*}

\begin{figure}[!t]
	\centering
	\includegraphics[width=0.99\columnwidth]{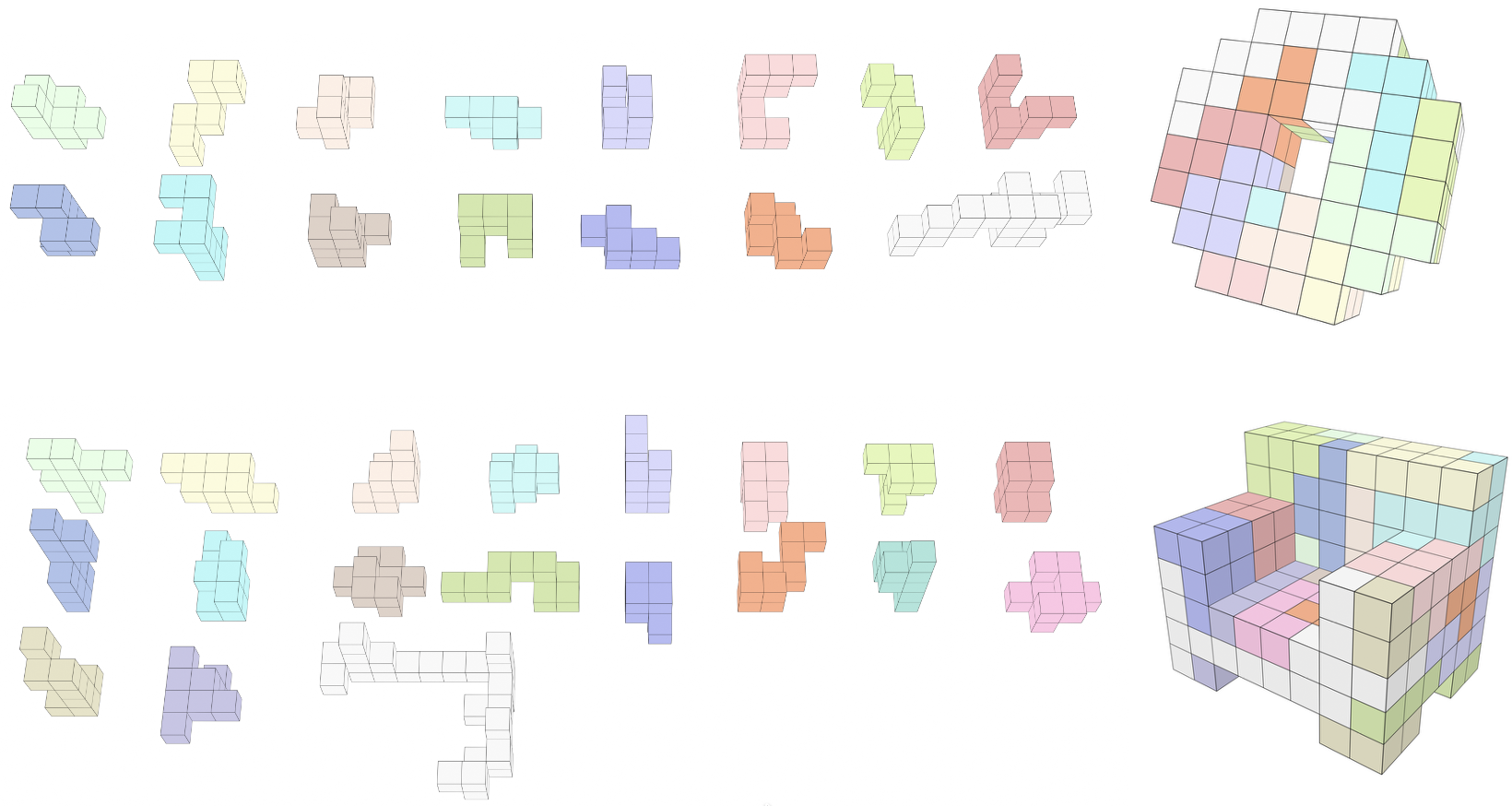}
	\caption{
		Representative interlocking assemblies with non-cubic shapes generated by RL-Lock:
		(top) a 15-piece Ring with resolution \(8 \times 8 \times 3\) and
		\(N=144\) voxels, and
		(bottom) a 19-piece Sofa with resolution \(8 \times 7 \times 6\) and
		\(N=208\) voxels.
		Each row shows the generated pieces on the left and the assembled shape
		on the right.
	}
	\label{fig:result_shapes}
\end{figure}

\paragraph{Comparison with DESIA.}
DESIA~\cite{Wang-2018-DESIA} is a heuristic-based search method that generates interlocking assemblies from a voxelized shape and a target piece count \(K\).
Different from DESIA, RL-Lock learns to generate interlocking assemblies from experience.


For this comparison, we reuse the nine generation tasks in the ablation study; see Table~\ref{tab:desia-comparison}.
RL-Lock follows the batched inference protocol described above.
For DESIA, we repeat independent generation runs until five valid solutions are obtained and report their mean generation time.
Each DESIA run is limited to 60 minutes; if no valid solution is found within this limit, the result is denoted by \textbackslash{}.
Table~\ref{tab:desia-comparison} shows that our RL framework can efficiently generate interlocking assemblies for challenging tasks in which DESIA requires excessive computation time (i.e., $(5^3, K=16)$, $(6^3, K=25)$) or fails to find a valid solution (i.e., $(4^3, K=9)$). 

Compared with DESIA, one limitation of our RL framework is that the neural network needs to be retrained when the user input changes, such as the input shape or the number of pieces $K$. 
Due to this reason, for a user who wants to generate one interlocking assembly, RL-Lock may be substantially more expensive due to the training cost. 
On the other hand, RL-Lock is more suitable for the case where a user generates many interlocking assemblies of the same class, which amortizes the training cost.
This experiment demonstrates that RL-Lock offers complementary strengths to DESIA, i.e., 1) excelling at solving challenging generation tasks and 2) allowing efficient generation of a large number of solutions of the same class.


\paragraph{Results.}
Figure~\ref{fig:result_cubes} visualizes three representative interlocking cube results generated by RL-Lock in the ablation study, where we show both the geometry of each piece as well as the resulting assembly.
Figure~\ref{fig:result_assembly_cube} further visualizes the assembly sequence of the 25-piece interlocking $6^3$ cube.

Beyond cubes, RL-Lock is able to generate interlocking assemblies for input shapes with irregular geometries.
Figure~\ref{fig:result_shapes} shows  two examples generated by RL-Lock, i.e., a Ring with a hole and a Sofa with non-convex features, demonstrating its ability to handle input shapes with different resolutions and geometric structures.
Figure~\ref{fig:result_assembly_sofa} further visualizes the assembly sequence of the 19-piece interlocking Sofa.
Please refer to the supplementary video for animations of the assembly processes of these results.

\section{Conclusion}
\label{sec:conclusion}

In this paper, we present a reinforcement learning framework RL-Lock for generating interlocking assemblies from voxelized 3D shapes. 
By formulating interlocking assembly generation as a sequential decision-making problem, our framework introduces action chunking to reduce the effective decision horizon and combines Monte Carlo tree search with a policy-value neural network to learn effective search strategies without relying on handcrafted heuristics.
Our experiments demonstrate that RL-Lock can efficiently generate interlocking assemblies and is particularly effective for challenging cases in which existing heuristic-based approaches struggle to find valid solutions. 
Our work demonstrates the potential of reinforcement learning to learn strategies for solving complex geometric design problems that traditionally rely on manually designed search heuristics, opening new directions for learning-based design of assemblies with complex geometric and/or structural constraints.

\bibliography{Ref_InterlockRL}

@STRING{TOG_SIG = {ACM Transactions on Graphics (Proc. of SIGGRAPH)}}

@STRING{TOG_SIGA = {ACM Transactions on Graphics (Proc. of SIGGRAPH Asia)}}

@STRING{CGF_EG = {Computer Graphics Forum (Proc. of Eurographics)}}

@STRING{CGF_PG = {Computer Graphics Forum (Proc. of Pacific Graphics)}}

@STRING{CAD = {Computer-Aided Design}}

@STRING{CAGD_GMP = {Computer Aided Geometric Design (Proc. of GMP)}}

@STRING{UIST = {Proc. of ACM UIST}}

@STRING{AAAI = {Proc. of the AAAI Conference on Artificial Intelligence}}

@STRING{NeurIPS = {Proc. of the Conference on Neural Information Processing Systems}}

@STRING{IROS = {Proc. of the IEEE/RSJ International Conference on Intelligent Robots and Systems}}

@inproceedings{Cicek-2016-3DUNet,
    author       =     {{\"{O}}zg{\"{u}}n {\c{C}}i{\c{c}}ek and Ahmed Abdulkadir and Soeren S. Lienkamp and Thomas Brox and Olaf Ronneberger},
	title           =      {3D U-Net: Learning Dense Volumetric Segmentation from Sparse Annotation},
    year           =     {2016},
	booktitle   =    {Medical Image Computing and Computer-Assisted Intervention (MICCAI)},
    pages        =    {424–432},
}

@software{Bradbury-2018-JAX,
	author       =     {James Bradbury and Roy Frostig and Peter Hawkins and Matthew James Johnson and Yash Katariya and Chris Leary and Dougal Maclaurin and George Necula and Adam Paszke and Jake Vander{P}las and Skye Wanderman-{M}ilne and Qiao Zhang},
	title           =     {{JAX}: Composable Transformations of {P}ython+{N}um{P}y Programs},
	url             =      {http://github.com/jax-ml/jax},
	year          =      {2018},
}

@inproceedings{Song-2022-InterlockingSurvey,
	author       =     {Peng Song},
	title           =    {Interlocking Assemblies: Applications and Methods},
	booktitle    =    {Materials Today: Proceedings (International Conference on Additive Manufacturing for a Better World)},
	volume     =    {70},
	year         =    {2022},
	pages      =    {78--82},
}

@book{Coffin-1990-Puzzling,
	author        =       {Coffin, Stewart T.},
	title            =        {The Puzzling World of Polyhedral Dissections},
	publisher   =        {Oxford University Press},
	year           =       {1990},
	pages        =        {196},
	ISBN         =        {0-19-286133-6},
}

@inproceedings{Sun-2024-StructCurves,
	author       =     {Zezhou Sun and Devin Balkcom and Emily Whiting},
	title            =    {{StructCurves}: Interlocking Block-Based Line Structures},
	booktitle    =    UIST,
	year          =    {2024},
	pages       =    {39:1--39:11},
}

@inproceedings{Zhang-2016-InterlockVoxel,
	title            =    {Interlocking Structure Assembly with Voxels},
	author        =    {Yinan Zhang and Devin Balkcom},
	booktitle     =    IROS,
	pages        =    {2173--2180},
	year           =    {2016},
}

@inproceedings{Shih-2016-SLBlocks,
	author       =    {Shen-Guan Shih},
	title           =    {On the Hierarchical Construction of {SL} Blocks},
	booktitle    =    {Proc. of Advances in Architectural Geometry},
	pages        =    {124--136},
	year          =    {2016},
}

@inproceedings{Wibranek-2021-SLBlocksAssembly,
	author      =    {Bastian Wibranek and Yuxi Liu and Niklas Funk and Boris Belousov and Jan Peters and Oliver Tessmann},
	title           =    {Reinforcement Learning for Sequential Assembly of SL-Blocks – Self-interlocking Combinatorial Design Based on Machine Learning},
	booktitle   =    {Proc. of the 39th Education and Research in Computer Aided Architectural Design in Europe (eCAADe) Conference},
	pages       =    {27--36},
	year         =    {2021},
}

@article{Chen-2022-HighLevelPuzzle,
	author     =    {Rulin Chen and Ziqi Wang and Peng Song and Bernd Bickel},
	title         =    {Computational Design of High-level Interlocking Puzzles},
	journal    =    TOG_SIG,
	volume   =    {41},
	number   =    {4},
	pages     =    {150:1--150:15},
	year       =    {2022}
}

@article{Wang-2019-TopoLock,
	author     =    {Ziqi Wang and Peng Song and Florin Isvoranu and Mark Pauly},
	title         =    {Design and Structural Optimization of Topological Interlocking Assemblies},
	journal    =    TOG_SIGA,
	year        =    {2019},
	volume    =    {38},
	number   =    {6},
	pages      =    {193:1--193:13},
}

@article{Tang-2019-3DDissection,
	author    =    {Keke Tang and Peng Song and Xiaofei Wang and Bailin Deng and Chi-Wing Fu and Ligang Liu},
	title         =    {Computational Design of Steady {3D} Dissection Puzzles},
	journal    =    CGF_EG,
	year        =    {2019},
	volume   =    {38},
	number  =    {2},
	pages     =    {291--303},
}

@article{Wang-2018-DESIA,
	author     =    {Ziqi Wang and Peng Song and Mark Pauly},
	title          =    {{DESIA}: A General Framework for Designing Interlocking Assemblies},
	journal    =    TOG_SIGA,
	year         =    {2018},
	volume   =    {37},
	number   =    {6},
	pages      =    {191:1--191:14},
}

@article{Song-2017-ReconfigInterlock,
	author     =    {Peng Song and Chi-Wing Fu and Yueming Jin and Hongfei Xu and Ligang Liu and Pheng-Ann Heng and Daniel Cohen-Or},
	title          =    {Reconfigurable Interlocking Furniture},
	journal    =    TOG_SIGA,
	volume    =    {36},
	number   =    {6},
	year         =    {2017},
	pages     =    {174:1--174:14},
}

@article{Yao-2017-InterlockShell,
	author      =    {Miaojun Yao and Zhili Chen and Weiwei Xu and Huamin Wang},
	title           =    {Modeling, Evaluation and Optimization of Interlocking Shell Pieces},
	journal     =    CGF_PG,
	volume    =    {36},
	number   =    {7},
	year         =    {2017},
	pages      =    {1--13},
}

@article{Fu-2015-Furniture,
	author      =    {Chi-Wing Fu and Peng Song and Xiaoqi Yan and Lee Wei Yang and Pradeep Kumar Jayaraman and Daniel Cohen-Or},
	title           =    {Computational Interlocking Furniture Assembly},
	journal      =    TOG_SIG,
	volume     =    {34},
	number    =    {4},
	year          =    {2015},
	pages       =    {91:1--91:11},
}

@article{Song-2015-Interlock,
	title           =    {Printing {3D} Objects with Interlocking Parts},
	author     =    {Peng Song and Zhongqi Fu and Ligang Liu and Chi-Wing Fu},
	journal     =    CAGD_GMP,
	volume    =    {35-36},
	pages      =    {137--148},
	year          =    {2015},
}

@article{Song-2012-InterCubes,
	author     =    {Peng Song and Chi-Wing Fu and Daniel Cohen-Or},
	title          =    {Recursive Interlocking Puzzles},
	journal    =    TOG_SIGA,
	year        =    {2012},
	volume   =    {31},
	number  =    {6},
	pages      =    {128:1--128:10},
}

@article{Xin-2011-BurrPuzzles,
	author      =    {Shi-Qing Xin and Chi-Fu Lai and Chi-Wing Fu and Tien-Tsin Wong and Ying He and Daniel Cohen-Or},
	title           =    {Making Burr Puzzles from {3D} Models},
	journal     =    TOG_SIG,
	year          =    {2011},
	volume    =    {30},
	number   =    {4},
	pages      =    {97:1--97:8},
}

@inproceedings{Li-2026-ReCAD,
	author        =     {Jiahao Li and Yusheng Luo and Yunzhong Lou and Xiangdong Zhou},
	title             =    {ReCAD: Reinforcement Learning Enhanced Parametric CAD Model Generation with Vision-Language Models},
	booktitle    =     AAAI,
	volume       =     {40},
	number      =     {8},
	pages         =     {6190--6198},
	year            =     {2026}
}

@inproceedings{Niu-2026-FromIntentToExecution,
	author       = 	   {Ke Niu and Haiyang Yu and Zhuofan Chen and Mengyang Zhao and Teng Fu and Bin Li and Xiangyang Xue},
	title            =     {From Intent to Execution: Multimodal Chain-of-Thought Reinforcement Learning for Precise CAD Code Generation},
	booktitle   =     AAAI,
	volume      =    {40},
	number     =    {10},
	pages        =    {8160--8167},
	year           =    {2026}
}

@inproceedings{Earle-2025-GameLevelDesign,
	author        =     {Sam Earle and Zehua Jiang and Eugene Vinitsky and Julian Togelius},
	title             =     {Video Game Level Design as a Multi-Agent Reinforcement Learning Problem},
	booktitle    =     {Proc. of the AAAI Conference on Artificial Intelligence and Interactive Digital Entertainment},
	pages         =     {32--42},
	year           =      {2025}
}

@inproceedings{Khalifa-2020-PCGRL,
	author        =      {Ahmed Khalifa and Philip Bontrager and Sam Earle and Julian Togelius},
	title             =     {PCGRL: Procedural Content Generation via Reinforcement Learning},
	booktitle    =      {Proc. of the AAAI Conference on Artificial Intelligence and Interactive Digital Entertainment},
	pages         =      {95--101},
	year            =      {2020}
}

@article{Li-2024-ProceduralMaterial,
	author         =       {Beichen Li and Yiwei Hu and Paul Guerrero and Milo{\v{s}} Ha{\v{s}}an and Liang Shi and Valentin Deschaintre and Wojciech Matusik},
	title              =       {Procedural Material Generation with Reinforcement Learning},
	journal         =       TOG_SIGA,
	volume        =       {43},
	number       =       {6},
	pages         =        {280:1--280:14},
	year            =        {2024}
}

@article{Choi-2025-CompliantMechanism,
	author        =     {Yejun Choi and Yeoneung Kim and Keun Park},
	title            =      {Deep Reinforcement Learning for Optimal Design of Compliant Mechanisms Based on Digitized Cell Structures},
	journal       =      {Engineering Applications of Artificial Intelligence},
	volume      =      {151},
	pages        =      {110702:1--110702:17},
	year           =      {2025}
}

@inproceedings{Whitman-2020-ModularRobot,
	author       =     {Julian Whitman and Raunaq Bhirangi and Matthew Travers and Howie Choset},
	title            =     {Modular Robot Design Synthesis with Deep Reinforcement Learning},
	booktitle   =     AAAI,
	volume     =     {34},
	number     =    {6},
	pages       =     {10418--10425},
	year          =     {2020}
}

@article{Liu-2026-AlphaZeroBlock,
	author       =     {Zhen Liu and Hongwei Deng and Yongcan Wang and Changzheng Yue and Tao Xie and Jiefeng Zhang and Chao Wu and Jiali Zhou and Zheyuan Zhou},
	title           =     {AlphaZeroBlock: A Deep Reinforcement Learning and Monte Carlo Tree Search Approach for Intelligent 3D Track Block Assembly},
	journal      =     {Expert Systems with Applications},
	volume      =     {324},
	pages        =     {Article No. 132439},
	year          =      {2026}
}

@article{Seriket-2025-Multi-materialPrinting,
	author       =     {Hichem Seriket and Oualid Bougzime and Yuyang Song and H. Jerry  Qi and Fr{\'e}d{\'e}ric Demoly},
	title            =     {Reinforcement Learning-enabled Design of Topological Interlocking Materials for Sustainable Multi-material Additive Manufacturing},
	journal       =     {Additive Manufacturing},
	volume      =     {111},
	pages        =     {104992:1--104992:12},
	year           =     {2025},
}

@inproceedings{Vallat-2023-Scaffold-freeConstruction,
	author       =     {Gabriel Vallat and Jingwen Wang and Anna Maddux and Maryam Kamgarpour and Stefana Parascho},
	title            =     {Reinforcement Learning for Scaffold-Free Construction of Spanning Structures},
	booktitle    =     {Proc. of the 8th ACM Symposium on Computational Fabrication},
	pages        =     {12:1--12:12},
	year          =     {2023}
}

@inproceedings{Chung-2021-BrickByBrick,
	author        =     {Hyunsoo Chung and Jungtaek Kim and Boris Knyazev and Jinhwi Lee and Graham W. Taylor and Jaesik Park and Minsu Cho},
	title             =     {Brick-by-Brick: Combinatorial Construction with Deep Reinforcement Learning},
	booktitle    =     NeurIPS,
	pages        =     {5745--5757},
	year           =     {2021}
}

@inproceedings{Hosmer-2020-SpatialAssembly,
	author       =    {Tyson Hosmer and Panagiotis Tigas and David Reeves and Ziming He},
	title           =    {Spatial Assembly with Self-Play Reinforcement Learning},
	booktitle    =    {ACADIA 2020: Distributed Proximities, Volume I: Technical Papers},
	pages        =    {382--393},
	year           =    {2020}
}

@article{Schulman-2017-PPO,
	title={Proximal policy optimization algorithms},
    author={Schulman, John and Wolski, Filip and Dhariwal, Prafulla and Radford, Alec and Klimov, Oleg},
    journal={arXiv preprint arXiv:1707.06347},
    year={2017}
}

@misc{Zhang-2024-SelfPlaySurvey,
	author            =      {Ruize Zhang and Zelai Xu and Chengdong Ma and Chao Yu and Wei-Wei Tu and Wenhao Tang and Shiyu Huang and Deheng Ye and Wenbo Ding and Yaodong Yang and Yu Wang},
	title                 =     {A Survey on Self-Play Methods in Reinforcement Learning},
	archiveprefix   =     {arXiv},
	eprint              =     {2408.01072},
	year                =     {2024}
}

@article{Silver-2016-AlphaGo,
	author        =     {David Silver and Aja Huang and Chris J. Maddison and Arthur Guez and Laurent Sifre and George van den Driessche and Julian Schrittwieser and Ioannis Antonoglou and Veda Panneershelvam and Marc Lanctot and Sander Dieleman and Dominik Grewe and John Nham and Nal Kalchbrenner and Ilya Sutskever and Timothy Lillicrap and Madeleine Leach and Koray Kavukcuoglu and Thore Graepel and Demis Hassabis},
	title            =     {Mastering the Game of Go with Deep Neural Networks and Tree Search},
	journal       =     {Nature},
	volume      =     {529},
	pages       =     {484--489},
	year          =     {2016}
}

@article{Silver-2017-AlphaGoZero,
	author        =     {David Silver and Julian Schrittwieser and Karen Simonyan and Ioannis Antonoglou and Aja Huang and Arthur Guez and Thomas Hubert and Lucas Baker and Matthew Lai and Adrian Bolton and Yutian Chen and Timothy Lillicrap and Fan Hui and Laurent Sifre and George van den Driessche and Thore Graepel and Demis Hassabis},
	title            =     {Mastering the Game of Go without Human Knowledge},
	journal       =     {Nature},
	volume      =     {550},
	pages       =     {354--359},
	year          =     {2017}
}

@article{Silver-2018-AlphaZero,
	author       =     {David Silver and Thomas Hubert and Julian Schrittwieser and Ioannis Antonoglou and Matthew Lai and Arthur Guez and Marc Lanctot and Laurent Sifre and Dharshan Kumaran and Thore Graepel and Timothy Lillicrap and Karen Simonyan and Demis Hassabis},
	title            =     {A General Reinforcement Learning Algorithm that Masters Chess, Shogi, and Go through Self-Play},
	journal       =     {Science},
	volume      =     {362},
	number     =     {6419},
	pages        =     {1140--1144},
	year          =     {2018}
}

@article{Wang-2025-LearningToAssemble,
	author       =     {Ziqi Wang and Wenjun Liu and Jingwen Wang and Gabriel Vallat and Fan Shi and Stefana Parascho and Maryam Kamgarpour},
	title           =     {Learning to Assemble with Alternative Plans},
	journal       =     TOG_SIG,
	volume      =     {44},
	number     =     {4},
	pages        =     {101:1--101:16},
	year           =     {2025}
}


\end{document}